\documentclass[10pt]{article}
\newif\ifblindreview

\usepackage[letterpaper]{geometry}
\usepackage{amta2024}
\usepackage{times}
\usepackage{url}
\usepackage{latexsym}
\usepackage{natbib}
\usepackage{layout}
\usepackage{multicol}
\usepackage{booktabs,array}
\usepackage{float}
\usepackage[hidelinks]{hyperref}
\usepackage[all]{hypcap}
\usepackage{graphicx}
\usepackage{inconsolata}
\usepackage{xcolor}

\usepackage{amsmath}
\usepackage{breqn}

\usepackage{enumitem}

\usepackage{tabularx}
\usepackage{booktabs}
\newcolumntype{R}{>{\centering\arraybackslash}X}
\usepackage{adjustbox}
\newcolumntype{Y}{>{\centering\arraybackslash}X}
\newcolumntype{L}{>{\centering\arraybackslash}m{3cm}}

\usepackage{multirow}

\newcolumntype{?}{!{\vrule width 1.5pt}}

\usepackage{times}
\usepackage[linesnumbered,ruled,vlined]{algorithm2e}

\hypersetup{
    colorlinks,
    linkcolor={red!50!black},
    citecolor={blue!50!black},
    urlcolor={blue!80!black}
}

\ifblindreview
  \newcommand{\authorinfo}{\author{}} 
\else
  \newcommand{\authorinfo}{
    \author{\name{\bf Hossam Amer*} \hfill  \addr{hossamamer@microsoft.com}
    \AND
            \name{\bf Abdelrahman Abouelenin*} \hfill \addr{ababouelenin@microsoft.com}
    \AND
           \name{\bf Mohamed Maher} \hfill \addr{momaherg@gmail.com}
    \AND
           \name{\bf Evram Narouz} \hfill \addr{evramnarouz@microsoft.com}
    \AND
           \name{\bf Mohamed Afify} \hfill \addr{mafify@microsoft.com}
    \AND
           \name{\bf Hany Awadallah} \hfill \addr{hanyh@microsoft.com}
    }
    
  }
\fi

\begin{document}

\amtaHeader{x}{x}{xxx-xxx}{2015}{45-character paper description goes here}{Author(s) initials and last name go here}
\title{Simply Trainable Nearest Neighbour Machine Translation with GPU Inference}

\authorinfo
\maketitle

\pagestyle{empty}

\begin{abstract}
\vspace{5pt}


Nearest neighbor machine translation is a successful approach for fast domain adaption, which interpolates the pre-trained transformers with domain-specific token-level k-nearest-neighbor (kNN) retrieval without retraining. Despite kNN MT's success, searching large reference corpus and fixed interpolation between the kNN and pre-trained model led to computational complexity and translation quality challenges. Among other papers, \cite{dai2023simple} proposed methods to obtain a small number of reference samples dynamically for which they introduced a distance-aware interpolation method using an equation that includes free parameters. This paper proposes a simply trainable nearest neighbor machine translation and carry out inference experiments on GPU. Similar to \cite{dai2023simple}, we first adaptively construct a small datastore for each input sentence. Second, we train a single-layer network for the interpolation coefficient between the knnMT and pre-trained result to automatically interpolate in different domains. Experimental results on different domains show that our proposed method 
either improves or sometimes maintain the translation quality of methods in \cite{dai2023simple} while being automatic. In addition, our GPU inference results demonstrate that knnMT can be integrated into GPUs with a drop of only 5\% in terms of speed.

\end{abstract}

\begin{multicols}{2}

\section{Introduction}

Neural Machine Translation (NMT) has been showing an increasing trend of translation quality owing to the ongoing development of deep neural network models \cite{vaswani2017attention, kim2021scalable}. However, the quality of these models is limited as soon as the domain of the input test sentences is different than the training data. 
\let\thefootnote\relax\footnote{* equal contribution}
To handle this out-of-domain problem, k-nearest neighbor machine translation (kNN-MT) has proven to be successful in many studies \cite{khandelwal2021nearest, zheng2021adaptive, zheng2021non, jiang2021learning, wang2022efficient, meng2022fast}, and thus piqued much attention in the community of machine translation. At the core of kNN-MT, a kNN classifier over an external datastore is built based on cached decoder representations and corresponding target tokens. This classifier is utilized to augment the given NMT model without finetuning leading to improved predictions, especially for domain adaption. Augmenting the NMT model is done via interpolating between the output probability distribution of the NMT model and the kNN classifier output probability distribution.

Despite kNN-MT's noticeable success in alleviating the domain adaption problem, vanilla kNN-MT proposed in \cite{khandelwal2021nearest} mainly suffers two challenges slowing down kNN-MT's deployment. First, vanilla kNN-MT requires large datastore sizes resulting in massive storage and expensive latency overheads during inference. For example, \cite{khandelwal2021nearest} showed that kNN-MT is two orders of magnitude slower than the base NMT system in a generation speed when retrieving 64 keys from a datastore containing billions of records. Second, the interpolation between the NMT model and the kNN classifier is fixed for all sentences in the test sets and manually tuned to improve translation quality.

Techniques in the literature have been proposed to overcome kNN-MT's challenges. For example, \cite{meng2022fast} designed Fast kNN-MT where a subset of the large datastore is created for each source sentence by searching for the nearest token-level neighbors of the source tokens and mapping them to the corresponding target tokens. Building on \cite{meng2022fast}, \cite{dai2023simple} proposed a simple and scalable kNN-MT that leverages current efficient text retrieval mechanisms, such as BM25 \cite{robertson2009probabilistic}, to obtain a small number of reference samples that have high semantic similarities with the input sentence, and then dynamically construct a tiny datastore by forwarding the samples to the pre-trained model. \cite{dai2023simple} successfully introduced a simple distance-aware interpolation equation to adaptively incorporate kNN retrieval results into the NMT model. However, this simple equation required manual tuning. Along the same line, \cite{jiang2022towards} proposed a trainable interpolation method but with six-layer neural network. To the best of our knowledge, these papers did not integrate kNN-MT into GPU inference to observe the trade-off between accuracy and speed results.

This paper proposes a simply trainable nearest neighbor machine translation via a single-layer neural network and demonstrates kNN feasibility with GPU inference. Similar to \cite{dai2023simple}, we reduce the large datastore size by extracting online a small number of reference samples that have high semantic similarities with the input test sentence using the efficient BM25 retrieval algorithm (\cite{robertson2009probabilistic}). Based on these insights, we propose a simply trainable single-layer neural network that adaptively interpolates the NMT and knnMT probability distributions per domain in an average of 40 minutes of a single GPU training time. Last but not least, we integrate kNN-MT into FasterTransformer, a highly optimized NMT GPU inference implementation offered by NVIDIA, and observe its speed and accuracy performance on a sparsely activated large-scale MoE. Experimental results show the translation quality effectiveness of our adaptive and automatic interpolation technique and insignificant speed drop of knnMT on GPU. Contributions of this paper are listed below:

\begin{itemize}[noitemsep,nolistsep]
    \item Propose and develop an adaptive and trainable single-layer neural network for knnMT interpolation that trains in 40 minutes. (Section \ref{sec:main_method}).
    \item Experimental results show that the proposed method improves the results of Sk-MT in some domains while maintaining the result in some other domains  (Section \ref{sec:trainable_results}).
    \item Integrate knnMT interpolation with GPU inference and introduce the speed-to-accuracy trade-off results. (Section \ref{sec:gpu_results}).
\end{itemize}

This remainder of this paper is organized as follows: Section \ref{sec:background} introduces knnMT and summarizes Sk-MT paving the road to the proposed method in Section \ref{sec:main_method}. Section \ref{sec:main_method} provides the experimental results while finally concluding in Section \ref{sec:conclusion}.

\section{Background: kNN-MT}
\label{sec:background}

\subsection{Vanilla-kNN}

In vanilla-kNN, a datastore is created to convert a bilingual sentence into a set of key-value pairs. These keys and values are defined in Equation \ref{eq:key_value_datastore}.

\begin{equation}
    K,V= F(x, y_{<t}),y_{t}
    \label{eq:key_value_datastore}
\end{equation}

where (x, y) $\in$ (X, Y) define the reference corpus for which the pretrained NMT model generates the context representation $F(x,y_{<t}$ at each time step $t$. Then we collect the output
hidden state $F(x,y_{<t})$ as key and $y_t$ as value to construct the whole datastore $(K, V)$.

At inference time, the current context representation $F(x, \hat{y}_{<t})$ at decoding step $t$, as well as the already generated words, are leveraged to generate a retrieval distribution $P_{knn}(y_{t}|y_{<t},x)$ over the entire vocabulary:

\begin{dmath}
  P_{knn}(y_{t}|x, \hat{y}_{<t}) \\\propto \sum_{(h_{i},v_{i})\in N_t}I_{y_t=v_i}  exp(\frac{-L_{2}(hi,F(x,\hat{y}_{<t}))}{T})
\label{eq:KNN_prob}
\end{dmath}

where L2 is the Euclidean distance between the current context embedding and the embedding of a token from the data store.
In vanilla KNN-MT, a predefined interpolation weight $\lambda$ is fixed as a hyperparameter. This weight interpolates between the probability distribution computed from KNN and the probability distribution generated from the pretrained NMT model (see Equation \ref{eq:interpolation}).

\begin{dmath}
   P(y_{t}|x, \hat{y}_{<t})= \lambda*P_{mt}( y_{t}|x, \hat{y}_{<t}) \\ +
    (1-\lambda)*P_{knn}(y_{t}|x, \hat{y}_{<t})
\label{eq:interpolation}
\end{dmath}

\subsection{SK-MT}


In SK-MT \cite{dai2023simple}, Elasticsearch is used for semantic retrieval components to create a sentence adaptive datastore instead of a static and extensive datastore used in Vanilla kNN-MT. In specific, Elasticsearch does two main operations: {Index} \& {Search}; storing parallel sentences in indexes format, and then retrieving 32 sentences per input sentence with the highest relevance score from the training corpus.

Also, SK-MT provided a successful way of setting the interpolation coefficient in Equation \ref{eq:adaptive_lambda}.

\begin{equation}
    \lambda=Relu(1-\frac{d_{0}}{T})
    \label{eq:adaptive_lambda}
\end{equation}

\noindent where $d_0$ is the top-1 L2 distance
during the nearest neighbor search, $T$ is the temperature parameter and is typically fixed.


\section{Single-Layer Trainable Interpolation}
\label{sec:main_method}


Even though SK-MT introduced a simple solution that derives the interpolation weight from the distance, a fixed parameter $T$ for all datasets is tuned to produce the best results and $T=100$ is recommended. A fixed temperature may not be optimal for all domains and datasets. For example, Table \ref{tab:koran_T} shows the BLEU score from the Koran dataset when varying $T$ from 100 to 500 with a step size of 100. As seen in the table, $T=300$ increases the BLEU score providing evidence that the temperature value can vary with the dataset and does not have to be the recommended SK-MT $T=100$. This observation motivates a simple and trainable method to adaptively find the temperature parameter for each dataset.

\begin{table}[H] 
\centering
\begin{tabular}{|c| c|}
\hline 
Temperature (T) &  BLEU
\\ \hline
100 &  15.5
\\ \hline
200 &  15.9
\\ \hline
300 &  16.1
\\ \hline
400 &  15.9
\\ \hline
500 &  15.5
\\ \hline
\end{tabular}
\caption{Koran Temperature Variation}
\label{tab:koran_T}
\end{table}


The proposed simple neural network consists of a single layer trained to predict the interpolation weight given the distance of the retrieved kNN candidates. This is in contrast to other adaptive interpolation methods e.g. \cite{jiang2022towards} that use more layers and learnable parameters. We use the development set of each domain to optimize our single-layer network.

Our training objective is designed to provide better translation quality. Knowing the ground truth token, we can choose the best interpolation weight that produces the best probability distribution that we can get from the interpolation between $P_{mt}$ and $P_{knn}$. Thus, our final objective is to create a sharper final probability distribution toward our ground truth token.

\begin{algorithm}[H]
\caption{Training Interpolation Layer}\label{alg:Training_procedure}
$len \gets length(y)$\;
$t \gets 0$\;
\While{$t \neq len$}{
    $gt \gets $ ground truth Index\;
    generate $Pknn(y_{<t}, x)$\;
    generate $Pmt(y_{<t}, x)$\;
    \eIf{$Pknn(gt|y_{<t}, x) \geq Pmt(gt|y_{<t}, x)$}{
        $label = 1$\; \tcp{favoring Pknn}
    }{
        $label = 0$\;
    }
    $\lambda_{pred} = Sigmoid(W \cdot D_{0} + B)$\;
    $loss \gets CrossEntropy(\lambda_{pred}, label)$\;
    update $W, B$\;
    $t = t + 1$\;
}
\end{algorithm}

As shown in Algorithm \ref{alg:Training_procedure}, our training procedure is divided into two stages at each decoding step.
The first stage examines the probability of the ground truth token in both distributions $P_{knn}$ and $P_{mt}$. If the probability of the ground truth token $P_{knn}$ is higher then we set the label to 1 otherwise we set the label to 0. The second stage trains our single-layer network using binary loss.



\begin{table*}[ht]
\caption{Translation quality of the proposed method versus other methods at Beam=5 and K=2.}
\begin{tabularx}{\textwidth}{?Y?Y?c?Y?c?Y?Y?}
\toprule
\specialrule{.1em}{.05em}{.05em}

\multicolumn{1}{ ?Y?}{\textbf{Domain}}  
& \multicolumn{3}{ c?}{\textbf{BLEU}}
& \multicolumn{3}{ c?}{\textbf{WMT22-COMET-da}}
\\
\cline{2-7}
& \multicolumn{1}{ c?}{\textbf{NMT}}
& \multicolumn{1}{ c?}{\textbf{SK-MT}}
& \multicolumn{1}{ c?}{\textbf{Trainable}}
& \multicolumn{1}{ c?}{\textbf{NMT}}
& \multicolumn{1}{ c?}{\textbf{SK-MT}}
& \multicolumn{1}{ c?}{\textbf{Trainable}}
\\
\cline{1-7}
IT & 38 & 45.5 & 46.1 & 83.0 & 85.0 & 85.0
\\
\cline{1-7}
Law & 49.6 & 62.8 & 62.7 & 86.7 & 88.3 & 88.0
\\
\cline{1-7}
Koran & 12.2 & 15.5 & 16.4 & 69.1 & 70.0 & 70.6
\\
\cline{1-7}
e-commerce & 52.5 & 58.1 & 58.5 & 90.7 & 90.9 & 90.9
\\
\cline{1-7}
finance & 48.6 & 53.3 & 53.3 & 70.6 & 94.2 & 93.9
\\
\cline{1-7}
\hline
\specialrule{.1em}{.05em}{.05em}
{Medical} & 42.7 & 57.1 &  57.2 & 83.9 & 85.2 & 85.0
\\
\hline
{medpharma} & 41.6 & 47.4 & 48.1 & 92.2 & 92.0 & 92.5
\\
\hline
\specialrule{.1em}{.05em}{.05em}
\textbf{AVERAGE} & 40.8 & 48.5 & 48.9 & 82.4 & 86.5 & 86.6
\\
\hline
\specialrule{.1em}{.05em}{.05em}
\end{tabularx}
\label{tab:main_results}
\centering
\end{table*}

\begin{table*}[ht]
\caption{GPU Inference Results on ZCode M3 Model.}
\begin{tabularx}{\textwidth}{?Y?l?Y?c?l?Y?c? }
\toprule 
\specialrule{.1em}{.05em}{.05em} 

\multicolumn{1}{ ?Y?}{} 
& \multicolumn{3}{ c?}{\textbf{beam=1, batch=1}}
& \multicolumn{3}{ c?}{\textbf{beam=2, batch=20}}
\\

\cline{2-7}

\textbf{Domain} & \multicolumn{2}{ c?}{\textbf{BLEU}}
& \multirow{2}*{\textbf{Speed Drop (\%)}}
& \multicolumn{2}{ c?}{\textbf{BLEU}}
& \multirow{2}*{\textbf{Speed Drop (\%)}}
\\

\cline{2-3}
\cline{5-6}
& NMT & knn-MT &  & NMT & knn-MT &
\\
\cline{1-7}
IT & 37.6 & 43.8 & 4.9 & 37.4 & 43.7 & 6.5
\\
\cline{1-7}
Medical & 45.6 & 55.6 & 5.0 & 45.8 & 56.3 & 9.1
\\
\cline{1-7}
Law & 54.1 & 61.8 & 5.8 & 54.1 & 62.2 & 5.5 
\\
\hline
\specialrule{.1em}{.05em}{.05em}
\textbf{AVERAGE} & \textbf{45.7} & \textbf{53.7} & \textbf{5.2} & \textbf{45.7} & \textbf{54.0} & \textbf{7.0}
\\
\hline
\specialrule{.1em}{.05em}{.05em}
\end{tabularx}
\label{tab:gpu_inference_results}
\centering
\end{table*}

\section{Experimental Results}
\label{sec:main_results}

\subsection{Experimental Setup}

\noindent \textbf{Input stimuli and Datasets:} We test our methodology in 2 language directions: German-English (deen), and English-Czech (encs). For deen, we employ the multi-domain dataset as the baseline \cite{khandelwal2021nearest} in addition to an e-commerce domain. For encs, we utilize two other domains: finance and medpharma. Our evaluation metrics are the SacreBLEU \cite{post-2018-call} and COMET-22 (wmt22-COMET-da) \cite{rei2022comet}, a reference-based metric that combines direct assessments (DA), sentence-level scores, and word-level tags from Multidimensional Quality Metrics (MQM) error annotations.



\noindent \textbf{Models:} Three transformer models are used in our experiments. The first two of the three are used to measure the translation quality; these two are constructed from 12 encoder layers and 12 decoder layers with 512 hidden dimensions and 2048 feedforward layer hidden dimensions with 8 multi-head attention heads. The third transformer is the ZCode M3 model reviewed and presented in \cite{kim2021scalable}. ZCode M3 is constructed from 24 encoder layers and 12 decoder layers with 1024 hidden dimensions and 4096 feedforward layer hidden dimensions with 16 multi-head attention heads. The ZCode M3 has 32 experts, 5B parameters, and 128,000 vocab size. 


\noindent \textbf{Baselines:} The model without knnMT is one baseline. To compare to other methods on our inhouse transformers, we utilize the SK-MT method that uses a distance-aware adapter \cite{dai2023simple}. In \cite{dai2023simple}, the authors compared with other methods and showed success so we use \cite{dai2023simple} as a transitive proxy to compare with other methods. 


\noindent \textbf{GPU Inference Hardware and Environment:} Inference and speed evaluation experiments are carried out on a single NVIDIA Tesla V100 GPU. Our inference environment is the highly optimized FasterTransformer from NVIDIA. Without loss of generality, we fix interpolation to $\lambda = Relu(1 - \frac{d_0}{100})$ and measure the speed.


\subsection{Trainable \textit{k}NN Retrieval Results}
\label{sec:trainable_results}

Table \ref{tab:main_results} shows the translation quality performance comparison between the proposed trainable method and other baselines. As shown in the table, our proposed trainable method improves the NMT baseline translation quality by a large margin. In addition, the proposed method 
improves or sometimes maintains the overall translation quality relative to SK-MT on average in terms of the BLEU and COMET 
scores. In some domains like IT and Koran, the proposed method improves the SK-MT performance. This result demonstrates the ability to at least maintain the performance of SK-MT while using a single-layer neural network. Also, these results overall show the adaptability of the proposed method to different datasets.  For Medical and medpharma, SK-MT outperforms our proposed method because the datastore built by the dev set does not have any semantic similarity to the training set leading to imbalanced binary labeling, whereas the test does not have this imbalanced binary labeling. To overcome this challenge, we suggest that we add weights to the binary cross-entropy training loss function. With this weighted loss function, our trainable method achieves 57.2 BLEU, 85 COMET in Medical, and 48.1 BLEU, 92.5 COMET in medphrama. These results increase our average results to 48.9 BLEU, and 86.6 COMET, respectively. 


Turning to the translation quality in terms of COMET, we observe that SK-MT still either maintain or improve the quality in the majority of the domains. We also notice that the improvements in BLEU scores do not fully transfer to COMET. We believe that this is due to the fact that COMET is trained on general domain data, therefore it's less sensitive to domain terminology and more focused on coherence and fluency. For example, e-commerce has an improvement of roughly 6 BLEU points relative to NMT, while the improvement is 0.3 COMET score points.

\subsection{GPU Inference Results}
\label{sec:gpu_results}



Table \ref{tab:gpu_inference_results} depicts the speed results of ZCode M3 inference and corresponding BLEU scores in three domains under test namely, IT, Medical, and Law. The kNN-MT results for beam=1, batch=1 setting on the large scale MoE improves the NMT baseline with a large margin while dropping the speed by only 5.2\% on average. Similarly, kNN-MT has an improved translation quality with only a drop of 7\% relative to NMT as beam and batch increase to 2 and 20, respectively. These results show the potential of deploying the knnMT domain adaption approach in such a large-scale model as ZCode M3.

\section{Conclusion}
\label{sec:conclusion}

This paper proposes a simply single-layer trainable nearest-neighbor machine translation and carries out experiments on large-scale models to demonstrate kNN feasibility with GPU Inference. Experimental results show the translation quality effectiveness of our adaptive and automatic interpolation technique relative to other methods in literature, the training simplicity in 40 mins on single-GPU,  and insignificant speed drop of knnMT on GPU inference.

\begin{small}
\bibliographystyle{apalike}
\bibliography{amta2024}
\end{small}




\end{multicols}
\end{document}